\documentclass{article}

\usepackage[preprint]{corl_2024} 

\usepackage{hyperref}
\usepackage{url}
\usepackage{amsmath} 
\usepackage{pgfplots}
\usepackage{pgfplotstable}
\usepackage[algo2e,algoruled,boxed,ruled]{algorithm2e}
\usepackage{graphicx}
\usepackage{subcaption}
\usepackage{tcolorbox}

\tcbuselibrary{listingsutf8}
\tcbuselibrary{skins}
\definecolor{lightgray}{gray}{0.9}
\usepackage{wrapfig}

\tcbuselibrary{breakable}

\usepackage{amsmath,amsfonts,bm}

\DeclareMathOperator{\E}{\mathbb{E}}

\title{Language-Model-Assisted Bi-Level Programming for Reward Learning from Internet Videos}

\author{
  Harsh Mahesheka\\
  Indian Institute of Technology (BHU) Varanasi\\
  \texttt{harsh.mahesheka.eee20@iitbhu.ac.in} \\ 
  \And
  Zhixian Xie \\
  Arizona State University \\
  \texttt{zxxie@asu.edu} \\
  \vspace{-10pt}
  \AND
  Zhaoran Wang \\
  Northwestern University \\
  \texttt{zhaoranwang@gmail.com} \\
  \And
  Wanxin Jin \\
  Arizona State University \\
  \texttt{wanxinjin@gmail.com} \\
}

\begin{document}
\maketitle

\vspace{-20pt}
\begin{abstract}
Learning from Demonstrations, particularly from biological experts like humans and animals, often encounters significant data acquisition challenges. While recent approaches leverage internet videos for learning, they require complex, task-specific pipelines to extract and retarget motion data for the agent. In this work, we introduce a language-model-assisted bi-level programming framework that enables a reinforcement learning agent to directly learn its reward from internet videos, bypassing dedicated data preparation. The framework includes two levels: an upper level where a vision-language model (VLM) provides feedback by comparing the learner’s behavior with expert videos, and a lower level where a large language model (LLM) translates this feedback into reward updates. The VLM and LLM collaborate within this bi-level framework, using a “chain rule” approach to derive a valid search direction for reward learning. We validate the method for reward learning from YouTube videos, and the results have shown that the proposed method enables efficiently reward design   from  expert videos  of biological agents for complex behavior synthesis. For videos showcasing our results and demonstrations used, please visit \href{https://drive.google.com/drive/folders/10OZKGuyQ70uWEPyEfWv6ihHSGuw6p2z1?usp=sharing}{\underline{videos}}.
\end{abstract}



\section{Introduction}
\vspace{-5pt}
Learning from demonstrations (LfD) offers an appealing way to enable robots or learner agents  to learn from  expert skills without expensive  programming from scratch \citep{scheel2022urban, silver2016mastering,ravichandar2020recent}. A learner agent can either  learn the expert policies via imitation learning \citep{chi2023diffusion, schaal1999imitation} or infer a reward function via inverse reinforcement learning \citep{ng2000algorithms} to derive its own  policies.
While  LfD \citep{schaal1999imitation,ng2000algorithms}  is early motivated from acquiring motion skills from biological agents like humans and animals, collecting ready-to-use expert data is always a challenge. Specialized setups, e.g., marker-based motion capture  \citep{de2009guide} and dedicated devices \citep{liu2024maniwav}, are often needed \citep{peng2020learning,zhang2018mode,kang2021animal}. The complexity of   data acquisition significantly limits the scalability of LfD methods, and many works \citep{peng2018deepmimic,holden2016deep} turn to animated motion data.

Methods for learning from internet videos \citep{zhang2023slomo,yao2022imitation,sivakumar2022robotic} offer promising solutions to data issue in LfD. However, video inputs adds another dimension of challenge in algorithm design. A dedicated preprocessing like pose estimation \citep{mathis2018deeplabcut,zhang2023slomo,yao2022imitation}  is typically needed to extract motion data from videos. Additionally,  motion retargeting is  sometimes required to address morphological  differences of biological experts or the differences of camera perspectives \citep{kim2022humanconquad,sivakumar2022robotic,bahl2022human}.

To address the data preparation challenges in learning from in-the-wild videos,  two questions need to be answered. \emph{Q1: Without data preparation, how can expert videos directly inform the improvement of robot behavior?} \emph{Q2: How to  represent robot behavior to facilitate the informed improvement?}

For Q1, rather than preprocessing videos, we record robot behavior as video clips and feed to a vision-language model (VLM), such as Google’s Gemini \citep{google_gemini} and OpenAI’s MiniGPT-4 \citep{openai_chatgpt}. By harnessing the semantic understanding and cross-modal reasoning capabilities of VLMs, we are able to directly obtain the semantic   feedback from VLM on how to improve robot behavior to emulate the expert's. Such feedback, while in textual form, can be used for robot motion improvement.

For Q2, we choose \emph{reward code} as a representation of robot behavior.  This is inspired by the recent success of 'language-to-reward' methods \citep{yu2023language,ma2023eureka,xietext2reward}. Thus, to update the reward given the feedback from VLM, we use a large language model (LLM) \citep{openai_chatgpt} to codify the textual feedback into the updates of executable  codes of robot reward. The updated reward will be optimized by the learner robot to derive its policy via reinforcement learning (RL).

The above VLM-LLM approach forms a bi-level framework for reward search, which we refer to as \emph{language-model-assisted bi-level programming}. It enables  to  a RL agent (robot) learns its reward directly from in-the-wild videos. The upper-level VLM provides textual feedback by the video clips of robot behavior with provided expert videos, and the lower-level LLM updates the robot reward based on the textual feedback. To our best knowledge, this is the first method to enable reward learning from in-the-wild videos without dedicated data preparations. It could pave the way toward a salable method for visual inverse reinforcement learning.

\section{Related Works}
\vspace{-5pt}
\paragraph{VLM/LLM in Reinforcement Learning and Robotics}
The near-human reasoning capabilities of LLMs make them well-suited for integration into agent decision making. For example, \citet{brohan2023can, lin2023text2motion, xie2023translating, singh2023progprompt, huang2024grounded} use LLM as high-level planners.  \citet{ xu2023creative, liu2024rl, liang2023code, ma2024explorllm} use LLMs as code generator. Other works use LLMs to guide skill acquisition \citep{liu2024rl},  evaluate environments \citep{xu2023creative}. LLMs have also been used  as exploration modules \citep{ma2024explorllm} and policy regularizer \citep{zhang2024can}.
VLMs further have the cross-modal understanding and reasoning abilities. They are thus  used to enhance semantic observations  \citep{chen2024vision}, extract language-conditioned features \citep{ma2024explorllm, jiang2024visual,  zeng2022socratic}. \citet{guan2024task,shridhar2022cliport, huang2024grounded} use VLMs in critiquing undesired behavior during the evaluation stages of RL.

\vspace{-5pt}
\paragraph{VLM/LLM for Reward Design}
Encoded with multi-domain knowledge, VLM/LLM can be used to  directly rate the quality of motion trajectories  \citep{zeng2023learning,wang2024rl}, generating reward signals \citep{kwon2023reward,baumli2023vision,rocamonde2023vision} in the RL training.  
 Recently, Eureka \citep{ma2023eureka} uses LLM  to refine the  reward coding. The method also shows  sim-to-real transfer with safety and domain-randomization  \citep{ma2024dreureka}.  Similarly, \citet{xietext2reward} proposes to use LLMs for few-shot reward code generation \cite{xietext2reward} with human feedback. \citet{yu2023language} use LLMs to generate the cost function code for model predictive control. While our proposed bi-level framework also involves LLM-assisted reward generation module, we focus on the problem of inverse reinforcement learning with a proposed higher-level VLM used to guide the reward search.

\vspace{-5pt}
\paragraph{Inverse Reinforcement Learning}
Our method  essentially falls to the category of inverse reinforcement learning (IRL) \citep{levine2011nonlinear,ziebart2008maximum,finn2016guided,ni2021f,xu2022receding}.  As noted earlier, a shared challenge of existing IRL methods is expert data acquisition. While recent efforts have explored learning rewards from  videos \citep{das2021model,kumar2023graph}, they typically require  task-specific video processing modules to extract expert motion.
In contrast, our method is preprocessing-free visual IRL. By leveraging the visual reasoning capabilities of foundational models, our approach directly updates the reward by intaking both expert and learner motion videos. This significantly improves the scalability of IRL for learning from the vast array of demonstration videos available on the internet.

\vspace{-5pt}
\paragraph{Visual Imitation Learning}
Complementary to our work are the methods that learn policies from videos of biological agents \citep{sermanet2016unsupervised,liu2018imitation}. A key is to extract and map the biological motion in videos to   robot embodiment. The features e.g., keypoints \citep{zhang2023slomo,yao2022imitation,sivakumar2022robotic,bahl2022human} or skeletons \citep{kim2022humanconquad} are typically extracted from videos. These features  will be used for re-target network training for teleoperation \citep{sivakumar2022robotic} or RL \citep{bahl2022human}. Extracting key features from videos typically involves heavy  engineering. Our method provides an alternative, yet direct way of learning rewards from video demonstrations.

\vspace{-5pt}
\paragraph{Learning from Biological Agents}
Works in \citep{peng2020learning,bohez2022imitate,yao2022imitation,zhang2023slomo} use imitation learning to transfer  animal skills into robots, but these methods are also engineering-heavy because motion capture and feature extractions are needed to obtain data. Compared with those biological imitation learning methods, our method is much less complex in implementation and provides a scalable solution to  learn  reward from animal videos for complex  behavior synthesis.


\section{Problem Statement}
We formulate a robot (learner agent)  as a Markov Process, defined by $\mathcal{M}=\langle \mathcal{S}, 
\mathcal{A},
\mathcal{T},
\rho_0
\rangle$.
Here, $\mathcal{S}$ and $\mathcal{A}$ denote the robot state and action space, respectively; $\mathcal{T}:\mathcal{S}\times\mathcal{S}\times\mathcal{A}\rightarrow[0,1]$ is the robot  dynamics; $\rho_0$ denotes the distribution of initial robot state. Given a  reward function $\mathcal{R}:\mathcal{S}\times\mathcal{A}\rightarrow \mathbb{R}$, the robot derives its optimal policy $\boldsymbol{a}_t\sim \pi_{\mathcal{R}}(\boldsymbol{a}_t | \boldsymbol{s}_t)$, by maximizing the expected reward
\begin{equation}\label{equ.rl_policy}
    \pi_{\mathcal{R}}=\arg \max_{\pi} \quad \E_{\pi, \mathcal{M}} \left[\sum\nolimits_{t=0}^{T}\gamma^t \mathcal{R}(\boldsymbol{s}_t, \boldsymbol{a}_t)\right],
\end{equation}
with  discount factor $\gamma\in(0,1)$. We define the \emph{robot behavior}   as the rollout trajectory of its optimal policy $\pi_{\mathcal{R}}$, i.e.,  
\begin{equation}\label{equ.rl_traj}
    \xi{(\pi_\mathcal{R})}=\{\boldsymbol{s}_0, \boldsymbol{a}_0, \boldsymbol{s}_1, ...., \boldsymbol{s}_{T-1}, \boldsymbol{s}_T\}\sim\rho_0 (s_0)\prod\nolimits_{t=0}^{T-1}\mathcal{T}(\boldsymbol{s}_{t+1}|\boldsymbol{s}_t, \boldsymbol{a}_t)\pi_\mathcal{R}(\boldsymbol{a}_t | \boldsymbol{s}_t).
\end{equation}

We follow \citep{jin2022learning} to formulate the problem of learning reward from video demonstrations. Consider the robot is given  expert  demonstration video $\mathcal{D}^e=\left\{{I}_{0:T}^e\right\}$, which is in an observation (video) space. We define an observation model $\mathcal{D}:\Xi\rightarrow\mathcal{D}$ to map the robot behavior $\xi{(\pi_\mathcal{R})}\in\Xi$ to the observation (video) space. 
\begin{equation}
    \mathcal{D}^r=\mathcal{D}\big(\xi{(\pi_\mathcal{R})}\big),
\end{equation}
which is a video of robot behavior $\xi{(\pi_\mathcal{R})}$.
Then, the problem of learning reward $\mathcal{R}$ from expert video $\mathcal{D}^e$ is 
\begin{equation}\label{equ.ldf_problem}
    \min_{\mathcal{R}}\quad \E_{\xi{(\pi_\mathcal{R})}}\,\,
    \mathcal{L}\Bigl(\mathcal{D}^e, \mathcal{D}\big(\xi{(\pi_\mathcal{R})}\big)\Bigr)
    \quad
    \text{with} \quad \pi_\mathcal{R} \,\, \,\, \text{is from } (\ref{equ.rl_policy}).
\end{equation}
Here,  $\mathcal{L}$ is a  \emph{imitation loss}, defined in the observation space, evaluating the mismatch between the expert motion and robot behavior in videos. The expectation is with respect to (\ref{equ.rl_traj}).

\subsection{Classic Gradient-Based Bi-Level Programming}
\vspace{-5pt}
LfD in (\ref{equ.ldf_problem}) is a bi-level optimization, where the upper level is to minimize visual imitation loss of the robot behavior $\xi{(\pi_\mathcal{R})}$ of policy $\pi_\mathcal{R}$, which is obtained from the lower-level through  reinforcement learning under the current reward $\mathcal{R}$. Classic methods, e.g., \citep{jin2022learning}, solve  (\ref{equ.ldf_problem}) based on  gradient-based methods, with the search direction of the reward given by 
\begin{equation}\label{equ.classic_bilevel}
\nabla_{\mathcal{R}}\mathcal{L}=\frac{\partial \mathcal{L}}{\partial\xi(\pi_{\mathcal{R}})}\frac{\partial\xi(\pi_{\mathcal{R}})}{\partial \mathcal{R}}.
\end{equation}
Each gradient-based update of the reward $\mathcal{R}$ requires solving the agent behavior $\xi(\pi_{\mathcal{R}})$ in the lower-level RL (\ref{equ.rl_policy}), and also differentiate through RL solution for gradient ${\partial\xi(\pi_{\mathcal{R}})}/{\partial \mathcal{R}}$.

Applying the classic bi-level programming (\ref{equ.classic_bilevel}) to visual LfD problem in (\ref{equ.ldf_problem}) faces one challenge: the visual loss $\mathcal{L}$ and observation model $\mathcal{D}$ has to be analytically given. This can be challenging for in-the-wild videos as it requires data-preprocessing and modeling effort. In the sequel, we will show how the capable VLMs and LLMs can be used as a gradient-free method to solve (\ref{equ.ldf_problem}).

\begin{figure}[h]
\begin{center}
\includegraphics[width=0.9\linewidth]{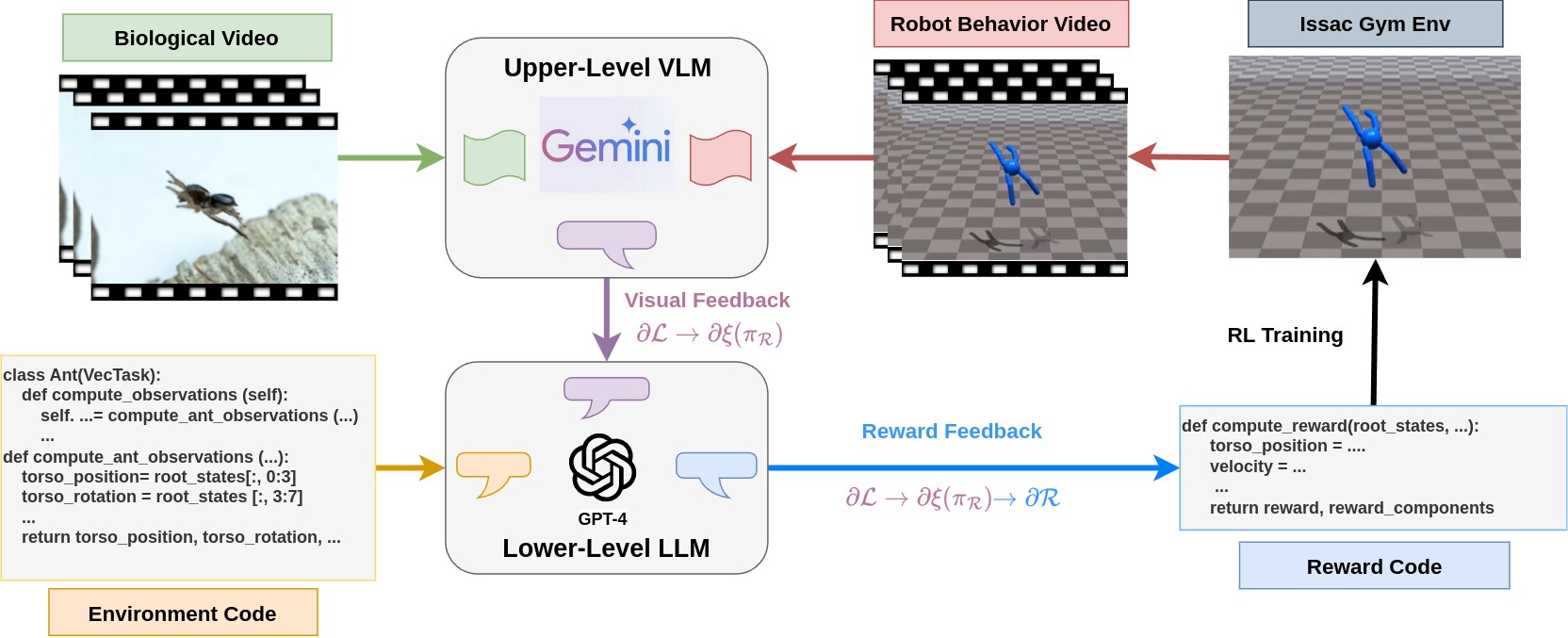}
\end{center}
\caption{Overview of the language-model-assisted bi-level framework for reward learning from videos. Upper-level VLM generates visual feedback by comparing the expert  video  and the video recording of  current robot behavior. Lower-level LLM uses the feedback to update the reward code. }
\label{fig.overview}
\end{figure}

\section{Language Model-Driven Bi-Level Programming}

\subsection{Framework Overview}
\vspace{-5pt}
A key insight of (\ref{equ.classic_bilevel}) is that the search direction of the reward $\mathcal{R}$ needs to combine two informed feedback via chain rule. (I) The first is   \emph{visual feedback} $\partial\mathcal{L}\rightarrow\partial\xi(\pi_{\mathcal{R}})$, from  visual imitation loss  to the robot behavior. This feedback informs how  to improve the robot behavior  to minimize visual imitation loss. (II) The second is \emph{reward update} $\partial\xi(\pi_{\mathcal{R}})\rightarrow\partial\mathcal{R}$, from the robot behavior $\partial\xi(\pi_{\mathcal{R}})$  to robot reward $\partial\mathcal{R}$. This informs how to update  reward in response to the  robot behavior improvement.

Based on the above insights, we propose the \emph{language model-driven bi-level programming} framework, as shown in Fig. \ref{fig.overview}. We use a VLM to generate the visual feedback $\partial\mathcal{L}\rightarrow\partial\xi(\pi_{\mathcal{R}})$, by showing VLM two videos: the expert demonstration video  $\mathcal{D}^e$ and a video recording of robot behavior $\mathcal{D}\big(\xi{(\pi_\mathcal{R})}\big)$. With the cross-modal understanding/reasoning capability, the VLM is able to provide the textual suggestions on how to improve robot  behavior to match the expert. Therefore, no specific modeling and video data processing is needed.

With the  visual feedback from VLM, the proposed framework uses a second LLM to instantiate the VLM's visual feedback (in the form of language) into the reward feedback, i.e., realizing  $\partial\xi(\pi_{\mathcal{R}})\rightarrow\partial\mathcal{R}$. This is similar to  \citep{yu2023language,xietext2reward,ma2023eureka}, which utilize the contextual and coding capabilities of LLM to achieve language-to-reward functionality. Since the robot reward function is represented as Pythonic codes, and thus LLM is able to directly update the code of robot reward  based on the textual prompt from VLM. The reward code is then used to derive  robot policy in an Issac Gym environment \citep{makoviychuk2021isaac}. In the sequel, we will provide the design of each components. 

\subsection{Upper-Level: Visual Feedback from VLM}
\vspace{-5pt}
Given the video clips of expert demonstration and a recording of robot's current behavior, the high-level VLM provides  feedback in the form of natural language about where the robot should improve its behavior to match  that of expert in the videos. Thus, the VLM plays the role of $\partial\mathcal{L}\rightarrow\partial\xi(\pi_{\mathcal{R}})$.
Notably, with minimal prompting, we find that VLM feedback is closely aligned with human feedback, indicating its strong ability to analyze and compare motion patterns across videos. Our experiment uses Gemini 1.5-Pro as our  VLM as it proved superior in video analysis during our experiments. An example of visual feedback is given in Fig. \ref{fig.feedback}. The complete set of prompts used is provided in the Appendix \ref{sec:High_Level_Prompt}.

\begin{figure}[h]
\begin{center}
\includegraphics[width=0.8\linewidth]{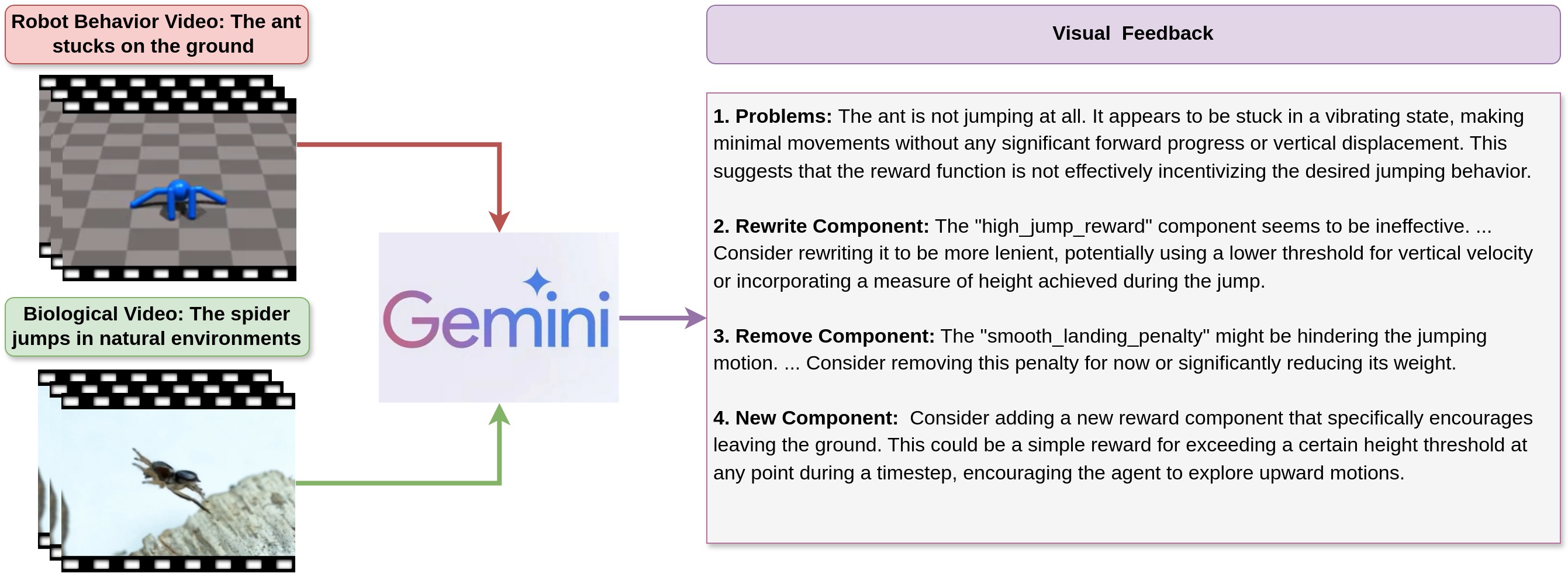}
\end{center}
\caption{One example of the upper-level visual feedback. Robot behavior recording and  biological (expert) motion video are provided to VLM, which then generates textual instructions for robot behavior improvement.}
\label{fig.feedback}
\end{figure}

\subsection{Lower-Level: Reward Feedback from LLM}
\vspace{-5pt}
The feedback from high-level VLM is then passed to the low-level LLM for reward update. The low-level LLM plays a role of translating the VLM's textual suggestion into the update of the robot reward. To achieve so,  similar to \cite{yu2023language,xietext2reward,ma2023eureka}, we represent the robot reward as code, which the LLM can directly manipulate due to is strong  code understanding and writing capabilities. Thus, the update of reward code from LLM is similar to the role of performing gradient  $\frac{\partial\xi(\pi_{\mathcal{R}})}{\partial\mathcal{R}}$. 

In our implementation, to generate effective reward update, we additionally pass  the robot environment code as a context to the low-level LLM, as in \citep{ma2023eureka}. This context reveals the necessary robot variables and environment parameters for successful code writing.  LLMs are capable of understanding large complex code, and thus no changes are needed in environment code. We also pass the values of current reward components at  intermediate policy training checkpoints. This helps ground the LLM reasoning and provides a reference for balancing the weights of reward components in its reward update. To help LLM codify the update of the reward, we pass some formatting and analysis tips as context. If the generated reward  code encounters errors during execution, the error messages are sent back to the LLM, which subsequently fixes the issues. We are using GPT-4o as our primary low-level LLM in our experimentation. As example of low-level LLM using  the high-level VLM's feedback to generate the update of the robot reward is shown in Fig. \ref{fig.reward_update}. The complete set of prompts used is provided in the Appendix \ref{sec:Low_Level_Prompt}.

\begin{figure}[h]
\begin{center}
\includegraphics[width=0.8\linewidth]{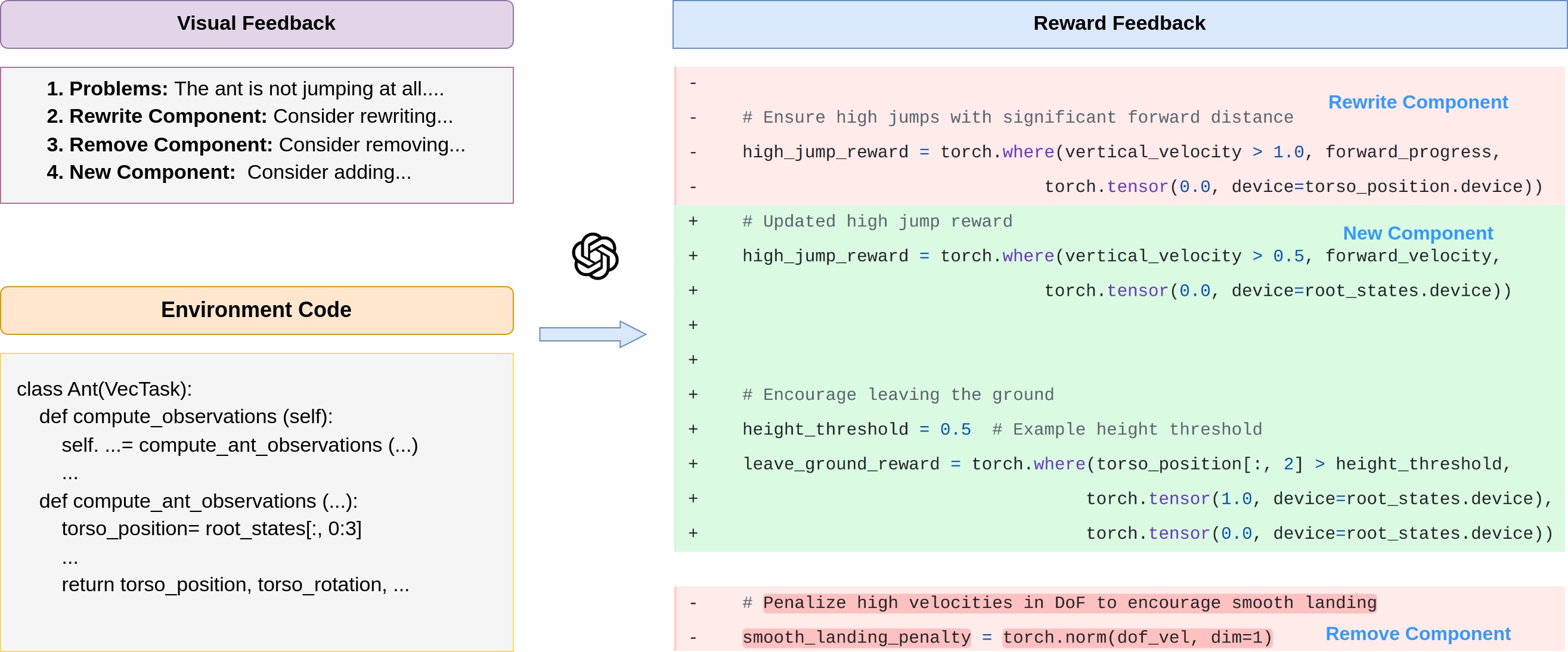}
\end{center}
\caption{One example of the reward code update for low-level LLM, by taking as input  the  language instructions from the VLM and environment code as context.}
\label{fig.reward_update}
\end{figure}

\begin{algorithm2e}
\setlength{\algomargin}{1.5em}
\small
    \caption{Language-Model-Assisted Bi-Level Reward Learning}
    \label{alg.overview}
    \SetKwInput{set}{Input}
    \set{\text{Number of iterations:} $I$, \text{Upper level VLM}, \text{Lower-level LLM}, \text{Environment:} $env$, \text{Expert video:} $\mathcal{D}^{e}$}
    \SetAlgoLined
    \SetKwInput{init}{Initialization}
    \init{VLM observes $\mathcal{D}^{e}$ and provides an initial description; \\
    \qquad\qquad\qquad LLM uses the  VLM's initial description to initialize the reward code $\mathcal{R}_0$;\\}
    \For{$i = 0$ \KwTo $I-1$}{
        Train a policy $\pi_{\mathcal{R}_i}$ in $env$ using PPO with the code of current reward $\mathcal{R}_i$;\\
        Roll out policy and get robot trajectory $\xi(\pi_{\mathcal{R}_i})$, and its video recording $\mathcal{D}(\xi(\pi_{\mathcal{R}_i}))$;\\
        \smallskip
        \textbf{Visual  feedback}: by shown  expert video   $\mathcal{D}^{e}$ and the robot video recording $\mathcal{D}(\xi(\pi_{\mathcal{R}_i}))$, 
        VLM generates textual feedback about how to improve robot behavior;\\

        \smallskip
        
        \textbf{Reward feedback}: with the feedback input from VLM, LLM generates the reward feedback by updating $\mathcal{R}_i$ to $\mathcal{R}_{i+1}$ in executable code;\\
              
    }
    \smallskip
    \Return{$\mathcal{R}_I$} 
\end{algorithm2e}

\section{Experiments}
\vspace{-5pt}

\begin{figure}[h]
    \centering
    
    \begin{subfigure}[b]{0.2\textwidth}
        \centering
        \includegraphics[height=1.4cm]
        {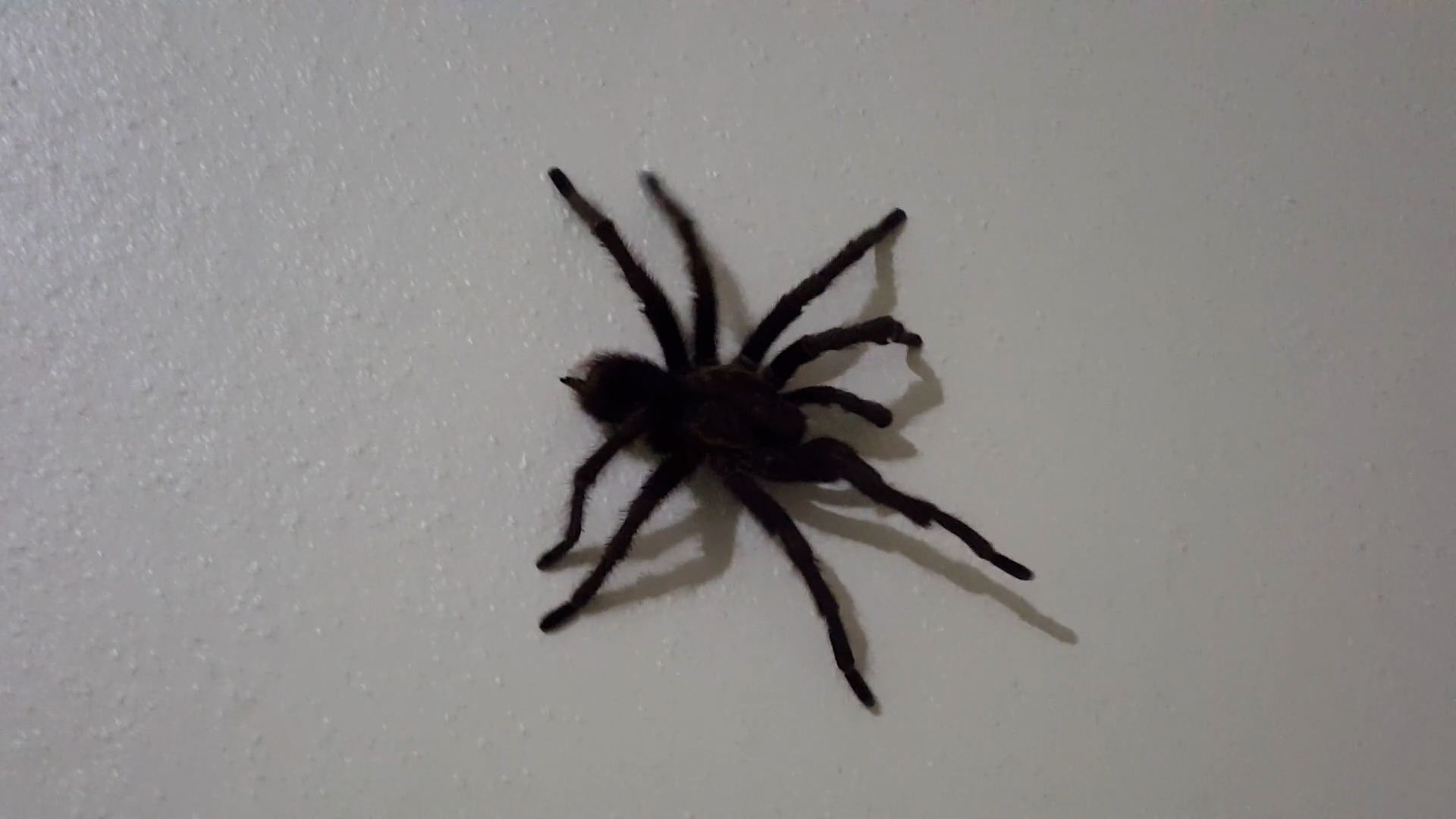}
        \caption{Spider Walking}
        \label{fig:video_subfig1}
    \end{subfigure}
    \begin{subfigure}[b]{0.2\textwidth}
        \centering
        \includegraphics[height=1.4cm]{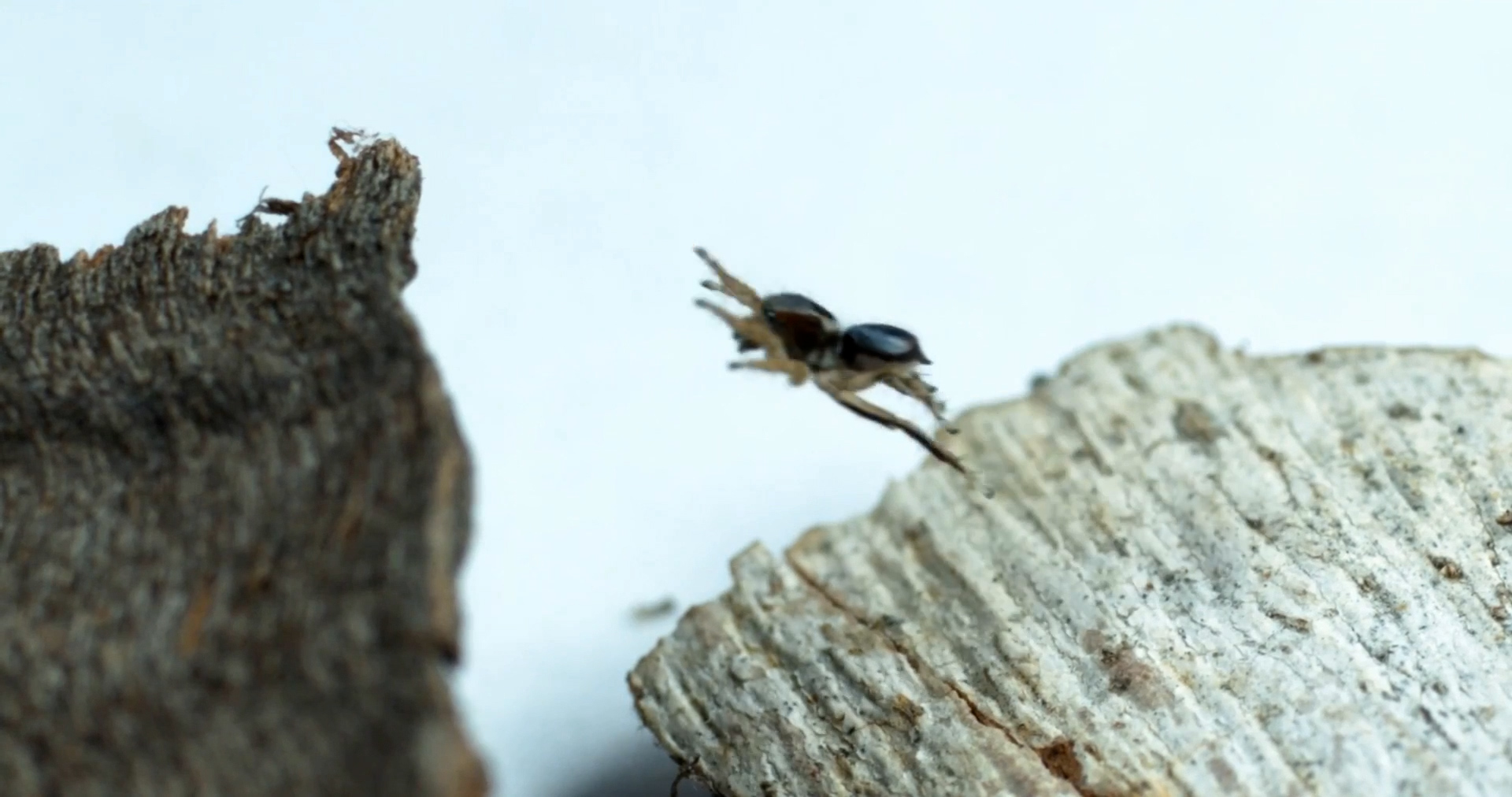}
        \caption{Spider Jumping}
        \label{fig:video_subfig2}
    \end{subfigure}
    \begin{subfigure}[b]{0.2\textwidth}
        \centering
        \includegraphics[height=1.4cm]{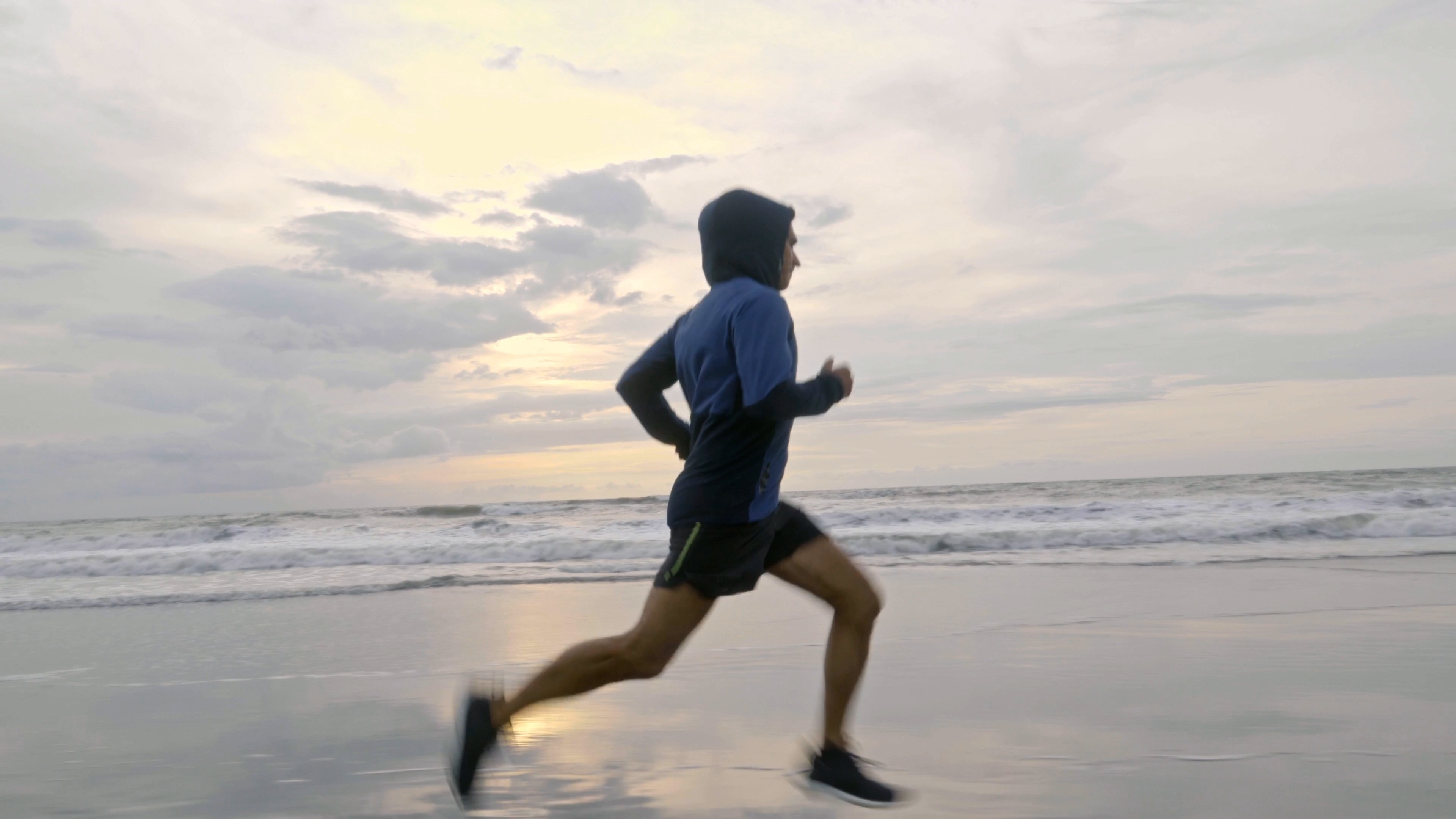}
        \caption{Human Running}
        \label{fig:video_subfig3}
    \end{subfigure}
    \begin{subfigure}[b]{0.2\textwidth}
        \centering
        \includegraphics[height=1.4cm]{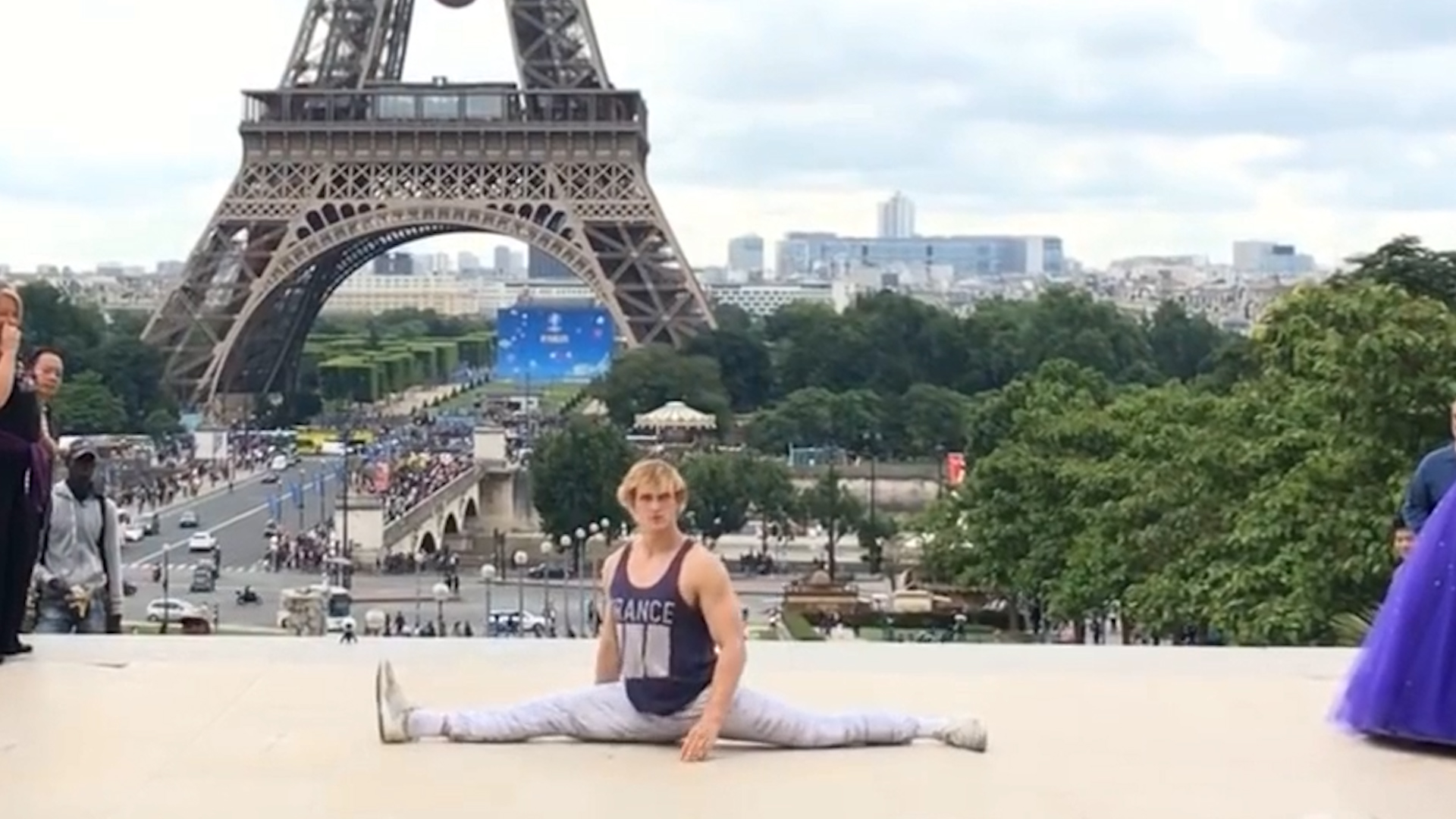}
        \caption{Human Splitting}
        \label{fig:video_subfig4}
    \end{subfigure}
    \begin{subfigure}[b]{0.16\textwidth}
        \centering
        \includegraphics[height=1.4cm]{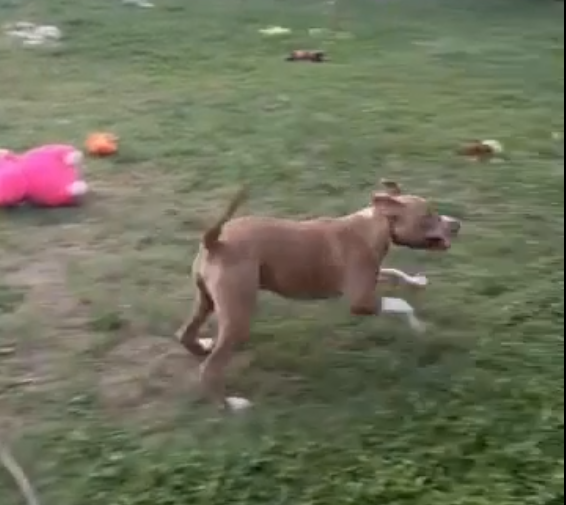}
        \caption{Dog Hopping}
        \label{fig:video_subfig5}
    \end{subfigure}
    \caption{Screenshots from YouTube videos used for reward learning}
    \label{fig:youtube_videos}
    \vspace{-5pt}
\end{figure}

We evaluate our approach on three robots—Ant, Humanoid, and ANYmal—in Isaac Gym environments \cite{makoviychuk2021isaac}, learning rewards from video demonstrations of their biological counterparts: spider, human athlete, and dog. The  biological motion videos, obtained directly from YouTube, are used to train the robots for various skillful motion tasks,  including \emph{Spider Walking},  \emph{Spider Jumping}, \emph{Human Running}, \emph{Human Split Landing},  \emph{Dog Hopping}, as shown in Fig. \ref{fig:youtube_videos}.






All  videos are from YouTube,  spanning 10-15 seconds \cite{pexels_man_running_2023, youtube_dog_hopping, youtube_human_splitting, youtube_spider_jumping, pexels_spider_crawling}. The selection of videos is flexible, and we only require the motion of a  biological agent is clearly presented in the video. To avoid confusion for the VLM, we avoid the videos where multiple agents are present.  We train the policies of the three robots, Ant, Humanoid, ANYmal in the Issac Gym environment, using  Proximal Policy Optimization (PPO) \citep{schulman2017proximal} with default hyperparameters. For each baseline, we run 5 independent runs and report the average performance achieved in these runs. Two example runs demonstrating the effectiveness of our approach are available in \ref{sec:Example_Runs}.  For videos showcasing our results and demonstrations used, please visit \href{https://drive.google.com/drive/folders/10OZKGuyQ70uWEPyEfWv6ihHSGuw6p2z1?usp=sharing}{\textbf{\underline{Videos}}}.

\subsection{Baselines}
\vspace{-5pt}
Given the limited baselines for reward learning from in-the-wild videos, we select baselines from the domains of LLM-assisted and human-feedback-based reward design. The maximum number of reward iteration for all methods is set to $I$ = 5.

\vspace{-5pt}
\textbf{Eureka}: a  state-of-the-art method that searches for RL reward  by LLM-assisted evolutionary optimization. It requires a fitness function for evaluation
of the agent policy. We follow the Eureka paper \citep{ma2023eureka} and set the fitness function (See Appendix \ref{sec:Appendix: Baseline Details}) that reflects the key biological motion features in the video for each learning task. The iteration batch size in Eureka is set $K=8$.

\vspace{-5pt}
\textbf{Eureka-GT}: We use a best-tuned score function (by an expert) as the fitness function in Eureka. As described later, this best-tuned  score  corresponds to the minimal  imitation loss for the biological motion in given videos.   We set the iteration batch size $K=4$ here.

\vspace{-5pt}
\textbf{Human as VLM}: This baseline is a modification of our bi-level method by replacing the high-level VLM with  a human to provide the visual feedback to the low-level LLM. This  is similar to Eureka with human input \citep{ma2023eureka} and text2reward \citep{xietext2reward}, except the goal here is learning from videos.

\subsection{Evaluation Metrics}
\vspace{-5pt}

\paragraph{Normalized Expert Score}
To evaluate the learned reward $\mathcal{R}$, we choose an expert best-tuned score function $\mathcal{S}_{\text{expert}}$ to evaluate the robot  behavior $\xi(\pi_{\mathcal{R}})$ of the reward $\mathcal{R}$. This score function is designed by human expert, through trial and error,  such that robot behavior $\xi(\pi_{\mathcal{S}_{\text{expert}}})$ generated by directly optimizing this best-tuned score  can capture the key biological motion features in the videos. 
 Thus,  a higher score value $\mathcal{S}_{\text{expert}}(\xi(\pi_{\mathcal{R}}))$ means  robot behavior is more similar to  the biological motion (this score can be considered as a negative visual imitation loss in (\ref{equ.ldf_problem})). The fitness functions used in \citep{ma2023eureka} are not suitable evaluation metrics because they are rudimentary, leading to agents exploiting the fitness criteria rather than genuinely learning the desired behaviors or motions.

To prove the validity of the best-tuned reward $\mathcal{S}_{\text{expert}}$, as will be shown later, we train  robot policy $\pi_{\mathcal{S}_{\text{expert}}}$ with it, and present the corresponding robot behavior $\xi(\pi_{\mathcal{S}_{\text{expert}}})$, together with the   robot behavior from the other learned rewards, to human ranking evaluation (next metric), and we find that the  $\xi(\pi_{\mathcal{S}_{\text{expert}}})$ is always ranked at top in the similarity to the biological motion video.


 In the performance report, we use normalized expert sore value, defined as  
\begin{equation}\label{equ.expert_score}
    S = {\mathcal{S}_{\text{expert}}(\xi({\pi}_{\mathcal{R}}))}/{\mathcal{S}_{\text{expert}}(\xi({\pi}_{\mathcal{S}_{\text{expert}}}))},
\end{equation}
where $\mathcal{S}_{\text{expert}}(\xi({\pi}_{\mathcal{R}}))$ is the  score value for the learned reward $\mathcal{R}$ by evaluating the corresponding robot behavior $\xi({\pi}_{\mathcal{R}})$ using the best-tuned score $\mathcal{S}_{\text{expert}}$.

\vspace{-5pt}
\paragraph{Human Preference Score}
We also conducted a user-study evaluation. The robot behavior recording $\mathcal{D}(\xi({\pi}_{\mathcal{R}}))$ from the learned reward $\mathcal{R}$, including $\mathcal{D}(\xi({\pi}_{\mathcal{S}_{\text{expert}}}))$ from best-tuned score, are presented to 10 participants for ranking. The participants were instructed to watch the video and score the robot behaviour on a scale from 0 to 5. Here,  0 indicates no correlation between the robot motion and the motion depicted in the video, while 5 signifies that the motions are very similar.

\subsection{Results and Analysis}
\begin{figure}[h]
\centering
\begin{minipage}{0.49\linewidth}
\centering
\includegraphics[width=\linewidth]{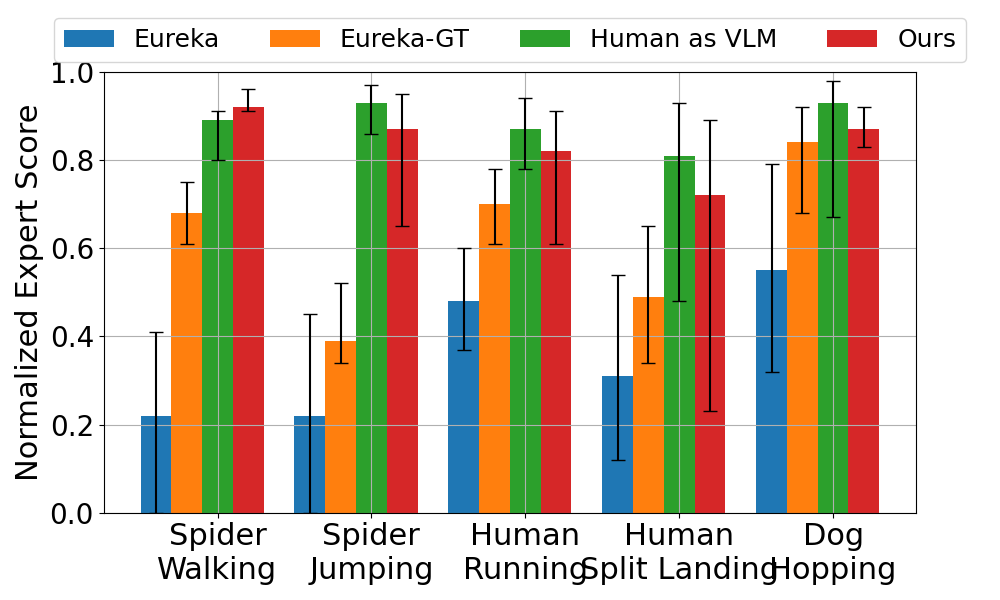}
\end{minipage}
\hfill
\begin{minipage}{0.49\linewidth}
\centering
\includegraphics[width=\linewidth]{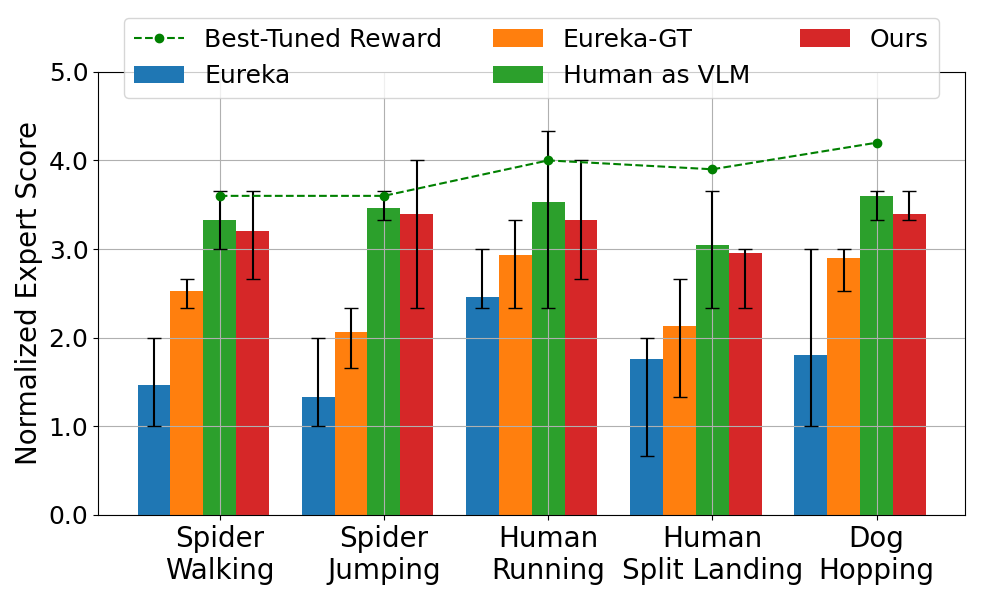}
\end{minipage}
\caption{(Left) Our approach consistently outperforms \emph{Eureka} and produces results similar to \emph{Human as VLM}, showcasing the ability of VLM to guide reward search in learning from biological traits. (Right) Human preferences further verify our claims that we can mimic complex traits from biological videos at par with an \emph{Human Expert}.}
\label{fig:combined_results}
\end{figure}

\vspace{-5pt}
Fig. \ref{fig:combined_results} shows the performance comparison between our method and three baselines (Eureka, Eureka-GT, Human-as-VLM) under two metrics (normalized expert score and human preference score), for the five LfD tasks (Fig. \ref{fig:youtube_videos}). All prompts to VLM and LLM for the tasks are given in Appendix \ref{sec:Full_prompts}.

\vspace{-5pt}
\textbf{Our approach outperforms Eurekas:} Fig. \ref{fig:combined_results} show that the proposed method consistently outperforms Eureka across all tasks. Comparison between Eureka and Eureka-GT shows the importance of choosing proper fitness function, which is typically difficult in LfD tasks or generally in reward design. Although Eureka-GT has the best possible fitness function, our approach outperforms Eureka-GT, proving the effectiveness of the reward search guided by VLM.

\vspace{-5pt}
\textbf{VLM feedback on par with human feedback:} Fig. \ref{fig:combined_results} shows a comparable performance between the proposed method and \emph{human as VLM}. Recall that the difference between the two methods is that we only replace  VLM with human in the high level. The results show VLM achieves near-human-level capability to provide meaningful feedback for robots to mimic biological counterparts.

\begin{wrapfigure}[11]{r}{0.5\linewidth} 
\vspace{-10pt}
    \centering
    \includegraphics[width=\linewidth]{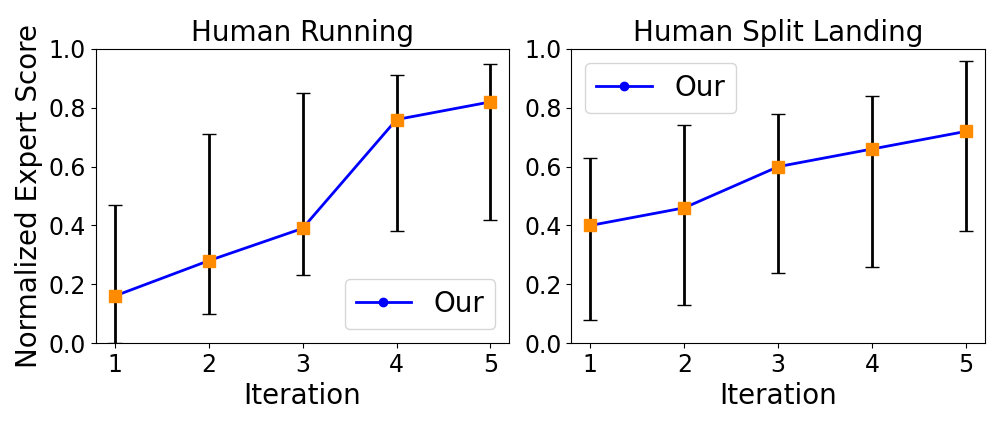}
    \vspace{-15pt}
    \caption{The normalized expert score after each iteration  of our method for two LfD tasks.}
    \label{fig.gradual_improv}
    \vspace{-2pt}
\end{wrapfigure}

\vspace{-5pt}
\textbf{Our method continuously improves reward:} Fig. \ref{fig.gradual_improv} shows the normalized expert score after each iteration during the reward update process of the proposed method for two LfD tasks. The continuous improvement of normalized expert score indicates that the search directions provided by the VLM and LLM are valid, and  the reward is effectively optimized to minimize the imitation loss.

\begin{figure}[h]
    \centering
    
    \begin{subfigure}[b]{0.99\textwidth}
        \centering
        \includegraphics[width=\textwidth]{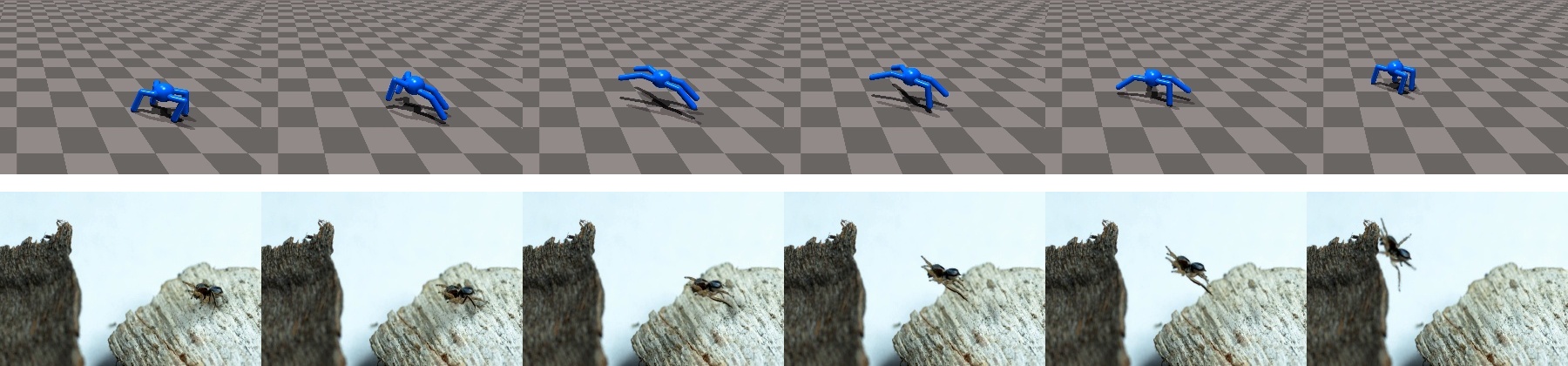}
        \caption{Spider Jumping}
        \label{fig:spider_sequence}
    \end{subfigure}
    
    \smallskip
    \begin{subfigure}[b]{0.99\textwidth}
        \centering
        \includegraphics[width=\textwidth]{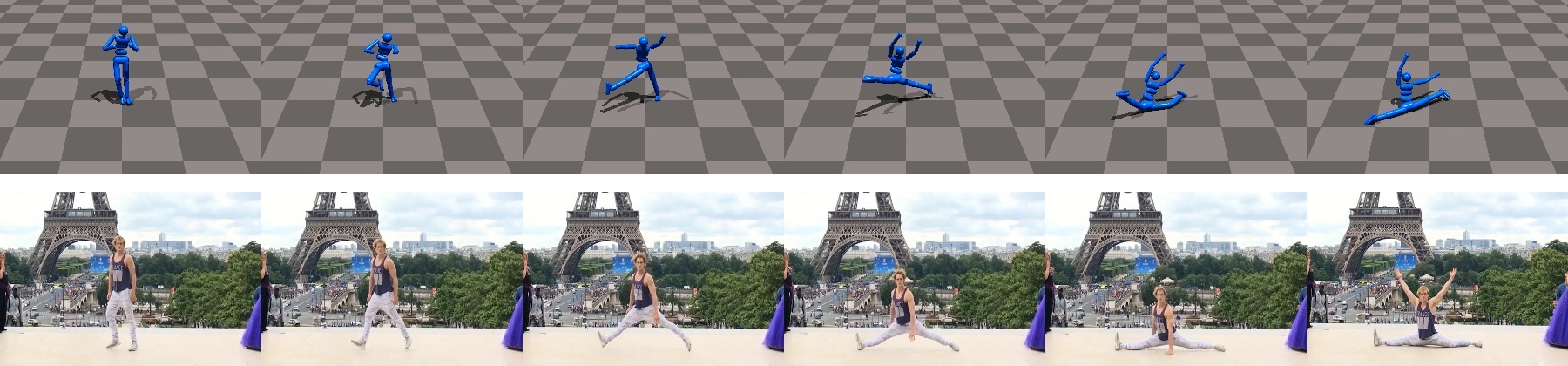}
        \caption{Human Split Landing}
        \label{fig:splitting_sequence}
    \end{subfigure}

    \caption{Robot behavior vs.  biological motion in the video for two tasks. }
    \label{fig:agent_behavior}
\end{figure}

\vspace{-5pt}
\textbf{The learned reward  captures complex motion skills in videos:} The learned reward by the our method is able to capture the fine-grained details of the biological motion in videos. For example, in Spider-Jumping task shown in Fig. \ref{fig:spider_sequence}, the Ant learns to crouch its body to a energy-storing position  before jumping, mimicking that of 
the spider. In the Human Split Landing in Fig. \ref{fig:splitting_sequence}, the Humanoid learns to have precise control of splitting both legs during descent, while coordinating upper body posture and arm movements to maintain stability. 
In Dog Jumpping, the Ant replicates the hopping and bouncing gait of a dog. These videos are included in the supplementary materials for further review, offering a visual demonstration of our method's capabilities.


\subsection{Ablation Study: Bi-Level versus Single-Level}

\begin{wrapfigure}[13]{r}{0.38\linewidth} 
\vspace{-40pt}
\begin{center}
\includegraphics[width=\linewidth]{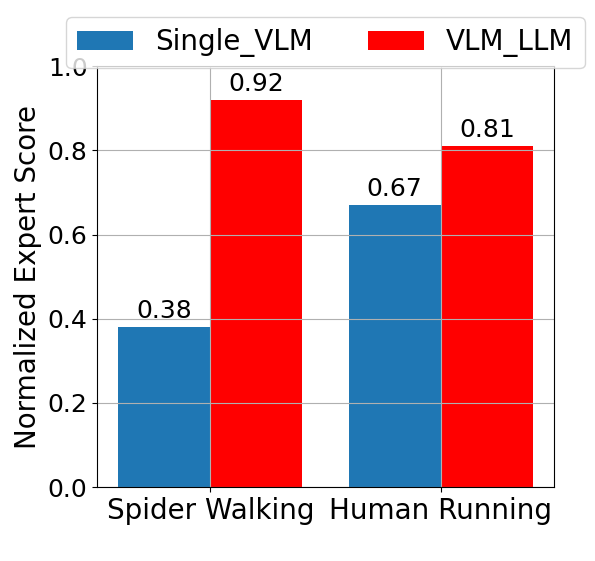}
\end{center}
\caption{Single VLM versus VLM-LLM bi-level (ours) methods.}
\label{fig.compare}
\end{wrapfigure}
To demonstrate the benefits of our bi-level design, we perform an ablation study to compare the proposed VLM-LLM  bi-level  method with single-level VLM that directly takes in expert videos and generates reward update. We perform five runs for the Spider Walking and Human Running tasks, and show comparison in Fig. \ref{fig.compare}. The results clearly show the advantage of the proposed  VLM-LLM design over the single-level VLM design. This advantage can be attributed to the hierarchical architecture, allowing VLM and LLM to specialize in their respective components: the VLM is tasked with high-level planning and the design of reward structures, while LLM is dedicated to environment-specific code generation and refinement. The results are consistent with prior research \cite{shinn2023reflexionlanguageagentsverbal, wang2024describeexplainplanselect}, where decomposing complex tasks into simpler, specialized components leads to better outcomes. 

\section{Conclusion}
This paper proposed a language-model-assisted bi-level programming framework to  learn reward functions from expert demonstration videos. The framework integrates a high-level VLM to analyze  expert  and learner behavior videos, generating visual feedback for the learner, and a low-level LLM to perform reward updates based on the VLM’s feedback. By leveraging the semantic and cross-modal  capabilities of the VLM, and the coding abilities of the LLM, the framework enables gradient-free reward search to minimize the visual imitation loss. We envision that this framework can extend existing inverse reinforcement learning methods, facilitating scalable learning from in-the-wild videos without requiring dedicated video processing routines.




\bibliography{mybib}  

\appendix

\section{Appendix}

\subsection{Full Prompts}
\label{sec:Full_prompts}
In this section, we present the complete prompts utilized by the high-level Visual Language Model (VLM) and the low-level Large Language Model (LLM). 
\subsubsection{High-Level VLM Prompts}
\label{sec:High_Level_Prompt}
\vspace{18pt}

\begin{tcolorbox}[colback=gray!10, colframe=black!80, sharp corners, boxrule=1pt, title=\textbf{High Level VLM: System Prompt}]
\footnotesize
\begin{verbatim}

This shows observation with all available variables and information about the
environment that LLM has. At any cost,limit the suggestion to what can be
implemented from this data:

{Task Obs Code String:}

\end{verbatim}

\end{tcolorbox}

\textbf{Task Obs Code String:} Raw environmental code provided to expose environmental specifications to LLM.

\vspace{18pt}
\begin{tcolorbox}[colback=gray!10, colframe=black!80, sharp corners, boxrule=1pt, title=\textbf{High Level VLM: Initial Prompt}]
\footnotesize
\begin{verbatim}
You are an expert in reinforcement learning and robotics. A Large Language Model
(LLM) is tasked with writing a reward function to train a reinforcement learning
agent to imitate the motion demonstrated by a {Creature Name} in the video.

The task is {Task Description}. Since the LLM does not have access to the video,
your task is to analyze the video and provide a detailed description of the
spider's jumping motion to assist the LLM in writing the reward function. The
description should contain the goal of the task along with necessary reward 
function considerations to help in its development. Do not include anything else.

The format of your response must be:

1. Task: A short description of the task.

2. Possible Reward Function Considerations (bullet points): List the three most
important components needed to make the task possible. Important: The components
should be easy to implement as the LLM writing reward has limited access to
information about the environment. 
\end{verbatim}

\end{tcolorbox}

\textbf{Creature Name:} Name of the biological agent you are trying to replicate.

\textbf{Task Description}: A short one-line description of the task you want the RL agent to pursue while mimicking the motion shown in the video.

\vspace{18pt}

\begin{tcolorbox}[colback=gray!10, colframe=black!80, sharp corners, boxrule=1pt, title=\textbf{High Level VLM: Review Prompt},     breakable,  fonttitle=\bfseries,  enhanced jigsaw ]
\footnotesize
\begin{verbatim}

I trained an agent and got the following results. Provide the major problems,
along with possible improvements in motion. The format of response should be :
1. Problems: Describe the motion and problems with by carefully studying the
video.
2. Rewrite Component: If some component seems to be wrongly implemented, 
suggest rewriting it.
3. Remove Component: If some component is not needed or is doing harm, then 
it is good to remove it.
4. New Component: Only if required, suggest new components needed; otherwise,
there are none. 

Tips for response: 
1. Study the videos, reward function, reward component values and prior
performances to provide feedback. 
2. Do not suggest complicated reward components which are hard to implement.
3. Keep the suggestion limited to what can be changed using the reward function.
4. The output should be only the 4 points mentioned above. 
Don't include anything else.
5. The Low-Level LLM writing reward function doesn't have access to video, so 
it cannot see or analyse motion in the video; instead, you should analyse it
and provide specific suggestions regarding the motion.

This was the reward function:
{Reward Function}
 
You can also use the values I tracked for the individual components in the 
reward function every {epochfreq} epoch and the maximum, mean, and minimum 
values encountered:

{Scalar values of all reward components}
\end{verbatim}

\end{tcolorbox}

\textbf{Reward Function:} The reward function is generated by low-level LLM.

\textbf{Scalar values of all reward components:} An automatically generated summary of the progress of various reward component values.

\vspace{18pt}

\subsubsection{Low-Level LLM Prompts}
\label{sec:Low_Level_Prompt}

\begin{tcolorbox}[colback=gray!10, colframe=black!80, sharp corners, boxrule=1pt, title=\textbf{Low Level LLM: System Prompt},     breakable,  fonttitle=\bfseries,  enhanced jigsaw ]
\footnotesize
\begin{verbatim}

You are a reward engineer trying to write reward functions to solve 
reinforcement learning tasks as effectively as possible. Your goal is to 
write a reward function for the environment that will help the agent learn
the task described in the text. Your reward function should use useful
variables from the environment as inputs. As an example, the reward function
signature can be: {Task Reward Signature String}

Since the reward function will be decorated with 
@torch.jit.script, please make sure that the code is compatible with TorchScript
(e.g.,use torch tensor instead of numpy array). 
Make sure any new tensor or variable you introduce is on the same device as
the input tensors. 
Please provide only the reward function without defining any new functions. 
Ensure that the code only uses existing PyTorch functions and do not introduce
any new functions.
\end{verbatim}

\end{tcolorbox}

\textbf{Task Reward Signature String}: A short example of the structure of the reward function required.

\vspace{18pt}

\begin{tcolorbox}[colback=gray!10, colframe=black!80, sharp corners, boxrule=1pt, title=\textbf{Low Level LLM: Initial Prompt},     breakable,  fonttitle=\bfseries,  enhanced jigsaw ]
\footnotesize
\begin{verbatim}
The Python environment is {Task Obs Code String}. Write a reward function for
the following task: {Task Description}.
The output of the reward function should consist of two items:
    (1) the total reward,
    (2) a dictionary of each individual reward component.
    
The code output should be formatted as a Python code string:
"```python ... ```".

Some helpful tips for writing the reward function code:
    (1) Make sure the type of each input variable is correctly specified; a 
    float input variable should not be specified as torch.Tensor
    (2) Try to keep the code and avoid writing overly complicated reward
    components.
    (3) Most importantly, the reward code's input variables must contain only
    attributes of the provided environment class definition (namely, variables
    that have the prefix self.).
    (4) Under no circumstance can you introduce new input variables.
\end{verbatim}
\end{tcolorbox}

\vspace{18pt}

\begin{tcolorbox}[colback=gray!10, colframe=black!80, sharp corners, boxrule=1pt, title=\textbf{Low Level LLM: Review Prompt},     breakable,  fonttitle=\bfseries,  enhanced jigsaw ]
\footnotesize
\begin{verbatim}
I trained an agent based on the reward function you provided and have the
following suggestions - 

{Suggestions from High-Level VLM}

For adjusting the weights you can use the values I tracked for the individual
components in the reward function every {epochfreq} epoch and the maximum, mean,
and minimum values encountered:

{Scalar values of all reward components}


Please prioritise addressing these problems and provide a new, improved reward
function that can better solve the task
\end{verbatim}

\end{tcolorbox}

\textbf{Suggestions from High-Level VLM:} Textual feedback from VLM

\vspace{18pt}

\begin{tcolorbox}[colback=gray!10, colframe=black!80, sharp corners, boxrule=1pt, title=\textbf{Low Level LLM: Error Resolving Prompt},     breakable,  fonttitle=\bfseries,  enhanced jigsaw ]
\footnotesize
\begin{verbatim}
Executing the reward function code above has the following error:
{Traceback Msg}. Please fix the bug and provide a new, improved reward function!
\end{verbatim}

\end{tcolorbox}

\textbf{Traceback Msg:} Error message generated while executing reward function 

\subsection{Baseline Details}
\label{sec:Appendix: Baseline Details}

The fitness functions used for running the Eureka baseline for each task are -

1. Spider Walking:
\[
F_{\text{SpiderWalking}} = \text{Forward\_Velocity}
\]

2. Spider Jumping:
\[
F_{\text{SpiderJumping}} = \text{Forward\_Velocity} + | \text{Vertical\_Velocity}| 
\]

3. Human Running:
\[
F_{\text{HumanRunning}} = \text{Forward\_Velocity}
\]

4. Human Splitting:
\[
F_{\text{HumanSplitting}} = \left| \text{Hip\_to\_right\_leg\_angle} - \frac{\pi}{2} \right| + \left| \text{Hip\_to\_left\_leg\_angle} - \frac{\pi}{2} \right| + |\text{Left\_knee}| + |\text{Right\_knee}|
\]

5. Dog Hopping:
\[
F_{\text{DogHopping}} = -(\text{linvel\_error} + \text{angvel\_error})
\]

The functions of tasks 1, 3, and 5 are directly derived from the Eureka paper, as these tasks closely align with those presented in the original work. The remaining tasks are extrapolated based on the same underlying principles and logic.
\subsection{Example Runs}
\label{sec:Example_Runs}

This section provides two comprehensive examples that demonstrate the functionality of our approach. Each example includes a detailed task description, the Environment Context provided, suggestions given by the VLM, and the corresponding reward function generated by LLM.

\subsubsection{Example 1: Spider Jumping}

\textbf{Task Description:} The task is to make ant move forward by performing high forward moving jumps on a flat simulated environment, imitating the spider's forward jumping style.
\vspace{20pt}
\begin{tcolorbox}[
    colback=gray!10, 
    colframe=black!80, 
    sharp corners,
    boxrule=1pt, 
    title=\textbf{Environment As Context}, 
    breakable, 
    fonttitle=\bfseries,
    enhanced jigsaw
]
\footnotesize
\begin{verbatim}


def compute_ant_observations(obs_buf, root_states, targets, potentials,
                             inv_start_rot, dof_pos, dof_vel, dof_limits_lower,
                             dof_limits_upper, dof_vel_scale, actions, dt,
                             basis_vec0, basis_vec1, up_axis_idx):

    torso_position = root_states[:, 0:3]
    torso_rotation = root_states[:, 3:7]
    velocity = root_states[:, 7:10]
    ang_velocity = root_states[:, 10:13]
    vertical_velocity = velocity[:, 2]
    forward_velocity = torch.norm(velocity[:, 0:2],
                                  p=2, dim=-1)

    to_target = targets - torso_position
    to_target[:, 2] = 0.0

    prev_potentials_new = potentials.clone()
    potentials = -torch.norm(to_target, p=2, dim=-1) / dt

    torso_quat, up_proj, heading_proj, up_vec, heading_vec = 
    compute_heading_and_up(
        torso_rotation, inv_start_rot, to_target, basis_vec0, basis_vec1, 2)

    vel_loc, angvel_loc, roll, pitch, yaw, angle_to_target = compute_rot(
        torso_quat, velocity, ang_velocity, targets, torso_position)

    dof_pos_scaled = unscale(dof_pos, dof_limits_lower, dof_limits_upper)

    obs = torch.cat((torso_position[:, up_axis_idx].view(-1, 1), vel_loc, 
                     angvel_loc, yaw.unsqueeze(-1), roll.unsqueeze(-1),
                     angle_to_target.unsqueeze(-1), up_proj.unsqueeze(-1),
                     heading_proj.unsqueeze(-1), dof_pos_scaled,
                     dof_vel * dof_vel_scale, actions), dim=-1)

    return obs, potentials, prev_potentials_new, up_vec, heading_vec
\end{verbatim}
\end{tcolorbox}
\vspace{18pt}
\begin{tcolorbox}[colback=gray!10, colframe=black!80, sharp corners, boxrule=1pt, title=\textbf{Iterative Loop},     breakable,  fonttitle=\bfseries,  enhanced jigsaw ]
\footnotesize
\begin{verbatim}
Iteration 0: Initialization

VLM Feedback 0:

1. Task: Train a four-legged simulated ant robot to move forward by performing
controlled high jumps, replicating the motion of a jumping spider. The jumps
should propel the ant forward while maintaining balance and stability for future
jumps.

2. Possible Reward Function Considerations:

* Reward high jump distance: Provide a large reward for maximizing the forward
distance covered by each jump. This encourages the agent to optimize its jumping
technique for maximum forward propulsion.

* Reward smooth landings: Penalize large impacts or instability on landing. This
encourages the agent to learn controlled landings, essential for balance and
future jumps.

* Reward consistent forward progress: Provide a reward for the agent's forward
progress over time. This encourages chaining successful jumps and avoids jumping
in place.

LLM Generated Code 0:

@torch.jit.script
def compute_reward(root_states: torch.Tensor, targets: torch.Tensor,
                   up_vec: torch.Tensor, dof_vel: torch.Tensor) -> 
                   Tuple[torch.Tensor, Dict[str, torch.Tensor]]:
    # Extract crucial states
    torso_position = root_states[:, 0:3]
    velocity = root_states[:, 7:10]
    vertical_velocity = velocity[:, 2]

    # Distance traveled forward
    forward_progress = velocity[:, 0] * 0.1

    # Ensure high jumps with forward distance
    high_jump_reward = torch.where(vertical_velocity > 1.0, forward_progress, 
                                   torch.tensor(0.0,
                                   device=torso_position.device))

    # Penalize high velocities in DoF for smooth landing
    smooth_landing_penalty = torch.norm(dof_vel, dim=1)

    # Reward for upright orientation
    upright_reward = up_vec[:, 2]

    # Combine rewards and penalties
    total_reward = (forward_progress + high_jump_reward + upright_reward
                    - smooth_landing_penalty)

    total_reward = torch.where(total_reward < 0, torch.tensor(0.0, 
                               device=torso_position.device), total_reward)

    reward_components = {
       "forward_progress": forward_progress,
       "high_jump_reward": high_jump_reward,
       "smooth_landing_penalty": smooth_landing_penalty,
       "upright_reward": upright_reward
    }
  
    return total_reward, reward_components
    
#################################################################

Iteration 1:

VLM Feedback 1:

1. Problems: The ant is not jumping at all. It seems stuck in a vibrating state
with minimal movements and no forward progress or vertical displacement.

2. Rewrite Component: The "high_jump_reward" seems ineffective. The condition
`vertical_velocity > 1.0` might be too strict. Consider lowering the threshold
or measuring height during the jump.

3. Remove Component: The "smooth_landing_penalty" may be hindering jumps. 
Without jumping, penalizing high DoF velocities may block necessary joint
movements. Consider removing or reducing its weight.

4. New Component: Add a reward for exceeding a height threshold during a 
timestep to encourage upward motion.

LLM Generated Code 1:

@torch.jit.script
def compute_reward(root_states: torch.Tensor, up_vec: torch.Tensor,
                  dof_vel: torch.Tensor) -> Tuple[torch.Tensor,
                  Dict[str, torch.Tensor]]:
    # Extract crucial states
    velocity = root_states[:, 7:10]
    torso_position = root_states[:, 0:3]
    forward_velocity = velocity[:, 0]
    vertical_velocity = velocity[:, 2]

    # Updated high jump reward
    high_jump_reward = torch.where(vertical_velocity > 0.5, forward_velocity, 
                                   torch.tensor(0.0, device=root_states.device))

    # Reward for leaving the ground
    height_threshold = 0.5
    leave_ground_reward = torch.where(torso_position[:, 2] > height_threshold, 
                                torch.tensor(1.0, device=root_states.device),
                                torch.tensor(0, device=root_states.device))

    # Forward progress reward
    forward_progress_reward = forward_velocity * 0.1

    # Reward for upright orientation
    upright_reward = up_vec[:, 2]

    # Combine rewards
    total_reward = (forward_progress_reward + high_jump_reward + 
                   leave_ground_reward + upright_reward)

    total_reward = torch.where(total_reward < 0, torch.tensor(0.0, 
                               device=root_states.device), total_reward)

    reward_components = {
       "forward_progress_reward": forward_progress_reward,
       "high_jump_reward": high_jump_reward,
       "leave_ground_reward": leave_ground_reward,
       "upright_reward": upright_reward
    }
  
    return total_reward, reward_components
    
#################################################################

Iteration 2:

VLM Feedback 2:

1. Problems: The ant is launching itself upward with excessive force, causing
uncontrolled flips and rotations.

2. Rewrite Component: The `high_jump_reward` is prioritizing upward velocity 
over forward velocity. Adjust it to encourage a forward trajectory.

3. Remove Component: The `leave_ground_reward` seems redundant. The
`high_jump_reward` already encourages the ant to leave the ground.

4. New Component: Add a penalty for excessive angular velocity during jumps to 
discourage uncontrolled flips.

LLM Generated Code 2:

@torch.jit.script
def compute_reward(root_states: torch.Tensor,
                  up_vec: torch.Tensor) -> Tuple[torch.Tensor,
                  Dict[str, torch.Tensor]]:
    # Extract crucial states
    velocity = root_states[:, 7:10]
    forward_velocity = velocity[:, 0]
    vertical_velocity = velocity[:, 2]
    ang_velocity = root_states[:, 10:13]

    # Reward for forward motion during jump
    high_jump_reward = forward_velocity + (0.5 * vertical_velocity)

    # Forward progress reward
    forward_progress_reward = forward_velocity * 0.1

    # Reward for upright orientation
    upright_reward = up_vec[:, 2]

    # Penalty for excessive angular velocity
    ang_velocity_mag = torch.norm(ang_velocity, p=2, dim=1)
    ang_velocity_penalty = -0.1 * ang_velocity_mag

    # Combine rewards
    total_reward = (forward_progress_reward + high_jump_reward 
                   + upright_reward + ang_velocity_penalty)

    total_reward = torch.where(total_reward < 0, torch.tensor(0.0, 
                               device=root_states.device), total_reward)

    reward_components = {
       "forward_progress_reward": forward_progress_reward,
       "high_jump_reward": high_jump_reward,
       "upright_reward": upright_reward,
       "ang_velocity_penalty": ang_velocity_penalty
    }
  
    return total_reward, reward_components

\end{verbatim}

\end{tcolorbox}

\subsubsection{Example 2: Human Running}
\textbf{Task Description:} The task is to make a humanoid run forward on a flat surface as fast as possible, replicating the gait shown in the video.
\vspace{20pt}
\begin{tcolorbox}[colback=gray!10, colframe=black!80, sharp corners, boxrule=1pt, title=\textbf{Evironment As Context},     breakable,  fonttitle=\bfseries,  enhanced jigsaw ]
\footnotesize
\begin{verbatim}
def compute_humanoid_observations(
    obs_buf, root_states, targets, potentials, inv_start_rot, dof_pos, dof_vel,
    dof_force, dof_limits_lower, dof_limits_upper, dof_vel_scale,
    sensor_force_torques, actions, dt, contact_force_scale, 
    angular_velocity_scale, basis_vec0, basis_vec1):

    torso_position = root_states[:, 0:3]
    torso_rotation = root_states[:, 3:7]
    velocity = root_states[:, 7:10]
    ang_velocity = root_states[:, 10:13]
    
    to_target = targets - torso_position
    to_target[:, 2] = 0

    prev_potentials_new = potentials.clone()
    potentials = -torch.norm(to_target, p=2, dim=-1) / dt

    torso_quat, up_proj, heading_proj, up_vec, heading_vec =
       compute_heading_and_up(torso_rotation, inv_start_rot, to_target,
                            basis_vec0, basis_vec1, 2)

    vel_loc, angvel_loc, roll, pitch, yaw, angle_to_target = 
       compute_rot(torso_quat, velocity, ang_velocity,  targets, torso_position)


    roll = normalize_angle(roll).unsqueeze(-1)
    yaw = normalize_angle(yaw).unsqueeze(-1)
    angle_to_target = normalize_angle(angle_to_target).unsqueeze(-1)
    
    dof_pos_scaled = unscale(dof_pos, dof_limits_lower, 
                             dof_limits_upper)

    obs = torch.cat((
        torso_position[:, 2].view(-1, 1), vel_loc, angvel_loc *
        angular_velocity_scale, yaw, roll,
        angle_to_target, up_proj.unsqueeze(-1), heading_proj.unsqueeze(-1), 
        dof_pos_scaled,
        dof_vel * dof_vel_scale, dof_force * contact_force_scale,
        sensor_force_torques.view(-1, 12) * contact_force_scale,
        actions), dim=-1)

    return obs, potentials, prev_potentials_new, up_vec, heading_vec
\end{verbatim}
\end{tcolorbox}

\vspace{20pt}
\begin{tcolorbox}[colback=gray!10, colframe=black!80, sharp corners, boxrule=1pt, title=\textbf{Iterative Loop},     breakable,  fonttitle=\bfseries,  enhanced jigsaw ]
\footnotesize
\begin{verbatim}
Iteration 0:

VLM Feedback 0:

1. Task: Make a humanoid run forward on a flat surface as 
fast as possible.

2. Possible Reward Function Considerations:
* Forward Velocity: Directly reward forward velocity. The faster the humanoid 
moves forward, the higher the reward.
* Stability Penalty: Discourage falling or unnatural movements with a penalty 
for large torso orientation changes or exceeding joint limits.
* Action Smoothness: Encourage smooth, human-like motion by penalizing large or 
sudden changes in joint actions. Penalize the derivative of the actions.

LLM Generated Code 0:

@torch.jit.script
def compute_reward(
   root_states: torch.Tensor,  dof_pos: torch.Tensor, dof_limits_lower: 
   torch.Tensor, dof_limits_upper: torch.Tensor, actions: torch.Tensor) -> 
   Tuple[torch.Tensor, Dict[str, torch.Tensor]]:

   # Calculate forward velocity (assuming forward is 
   # along the x-axis)
   forward_velocity = velocity[:, 0]

   # Reward for forward velocity
   forward_velocity_reward = forward_velocity

   # Penalty for large orientation changes (assuming 
   # stable when z-component is close to 1)
   torso_orientation_z = root_states[:, 2]
   stability_penalty = -torch.abs(torso_orientation_z - 1.0)

   # Penalty for exceeding joint limits
   dof_pos_scaled = (dof_pos - dof_limits_lower) / 
   (dof_limits_upper - dof_limits_lower)
   
   joint_limit_penalty = torch.sum(torch.where(dof_pos_scaled  > 1.0, 
   dof_pos_scaled - 1.0, torch.zeros_like(dof_pos_scaled)), dim=-1)
   
   joint_limit_penalty += torch.sum(torch.where(dof_pos_scaled < 0.0,
   -dof_pos_scaled, torch.zeros_like(dof_pos_scaled)), dim=-1)

   # Penalty for action smoothness (using absolute actions as a proxy)
   action_smoothness_penalty = torch.sum(torch.abs(actions), dim=-1)
  
   # Total reward
   reward = forward_velocity_reward + stability_penalty 
   reward -= joint_limit_penalty + action_smoothness_penalty

   # Return total reward and reward components
   reward_components = {
       "forward_velocity_reward": forward_velocity_reward,
       "stability_penalty": stability_penalty,
       "joint_limit_penalty": joint_limit_penalty,
       "action_smoothness_penalty": action_smoothness_penalty
   }
  
   return reward, reward_components

#################################################################

VLM Feedback 1:

1. Problems: The humanoid demonstrates an unnatural gait that lacks human 
running characteristics. It exhibits shuffling behavior where feet stay in 
contact with the ground. The humanoid also displays excessive arm movement,
swinging them unrealistically high.

2. Rewrite Component: stability_penalty: The current penalty may overly penalize
natural forward lean during running. Focus on penalizing large torso rotations 
along axes indicating instability.

3. Remove Component: joint_limit_penalty: This penalty seems ineffective as the
humanoid rarely approaches its joint limits. It can be removed.

4. New Component: Posture Reward: Introduce a reward to encourage a more upright
torso posture. The current motion is excessively bent forward. This could
involve rewarding torso height or penalizing forward tilt.

LLM Generated Code 1:

@torch.jit.script
def compute_reward(root_states: torch.Tensor, up_vec: torch.Tensor, actions:
    torch.Tensor) -> Tuple[torch.Tensor, Dict[str, torch.Tensor]]:
    
   # Extract torso position (first 3 components) and velocity (7-10) from 
   root_states
   torso_position = root_states[:, 0:3]
   velocity = root_states[:, 7:10]
   ang_velocity = root_states[:, 10:13]

   # Calculate forward velocity (assuming forward is along the x-axis)
   forward_velocity = velocity[:, 0]

   # Reward for forward velocity
   forward_velocity_reward = forward_velocity * 10.0

   # Penalty for torso roll and pitch deviations (focusing on stability)
   roll = ang_velocity[:, 0]
   pitch = ang_velocity[:, 1]
   stability_penalty = -torch.abs(roll) - torch.abs(pitch)

   # Posture reward to encourage upright torso height, penalizing forward tilt
   upright_torso_height = torso_position[:, 2]
   ideal_height = 1.0  # assuming 1.0 is ideal height
   height_penalty = -torch.abs(upright_torso_height - ideal_height)

   # Penalty for unnatural arm movements, using z-component of up_vec
   arm_movement_penalty = -torch.sum(torch.abs(up_vec[:, 2] - 
   1.0) * 0.1, dim=-1)

   # Penalty for action smoothness, using absolute actions
   action_smoothness_penalty = -torch.sum(torch.abs(actions) * 0.1, dim=-1)
  
   # Total reward
   reward = (forward_velocity_reward + stability_penalty + 
             height_penalty + action_smoothness_penalty + 
             arm_movement_penalty)

   # Return total reward and reward components
   reward_components = {
       "forward_velocity_reward": forward_velocity_reward,
       "stability_penalty": stability_penalty,
       "height_penalty": height_penalty,
       "action_smoothness_penalty": action_smoothness_penalty,
       "arm_movement_penalty": arm_movement_penalty
   }
  
   return reward, reward_components

#################################################################

VLM Feedback 2:

1. Problems: The agent is not running. It shows a strange shuffling motion with
outstretched arms. The torso is too upright and doesn’t lean forward like a 
natural runner.

2. Rewrite Component: arm_movement_penalty: This penalty encourages unnatural
arm behavior. Rewrite it to penalize arm angular velocity to keep the arms still.

3. Remove Component: height_penalty: This component is not contributing 
positively and could interfere with the learning of a natural forward lean
during running.

4. New Component: Torso Pitch Reward: Add a component to reward a slight forward 
lean. Reward the torso's pitch angle within a natural range for running.

LLM Generated Code 2:

@torch.jit.script
def compute_reward(
   root_states: torch.Tensor, up_vec: torch.Tensor, actions: torch.Tensor) -> 
   Tuple[torch.Tensor, Dict[str, torch.Tensor]]:
   
   # Extract torso position (first 3 components) and velocity (7-10) from 
   root_states
   torso_position = root_states[:, 0:3]
   velocity = root_states[:, 7:10]
   ang_velocity = root_states[:, 10:13]

   # Calculate forward velocity (assuming forward is along the x-axis)
   forward_velocity = velocity[:, 0]

   # Reward for forward velocity 
   forward_velocity_reward = forward_velocity * 2.0

   # Penalty for torso roll and pitch deviations (focusing on stability)
   roll = ang_velocity[:, 0]
   pitch = ang_velocity[:, 1]
   stability_penalty = -(torch.abs(roll) + torch.abs(pitch)) * 0.1

   # Remove height penalty and Penalize arm angular velocity to keep arms still
   arm_angular_velocity = ang_velocity[:, 1:3]
   arm_movement_penalty = -torch.sum(torch.abs(arm_angular_velocity), 
   dim=-1) * 0.5

   # Penalize sudden changes in actions by penalizing action values
   action_smoothness_penalty = -torch.sum(actions ** 2, 
   dim=-1) * 10.0

   # Reward for forward lean (positive pitch indicates forward lean)
   torso_pitch = ang_velocity[:, 1]
   lean_reward = torch.clamp(torso_pitch, -0.1, 0.1) * 2.0

   # Total reward
   reward = (forward_velocity_reward + stability_penalty + 
             action_smoothness_penalty + arm_movement_penalty 
             + lean_reward)

   # Return total reward and reward components
   reward_components = {
       "forward_velocity_reward": forward_velocity_reward,
       "stability_penalty": stability_penalty,
       "action_smoothness_penalty": action_smoothness_penalty,
       "arm_movement_penalty": arm_movement_penalty,
       "lean_reward": lean_reward
   }
  
   return reward, reward_components

\end{verbatim}
\end{tcolorbox}

\end{document}